\documentclass{article}

\usepackage[main, final]{neurips_2026}

\usepackage[utf8]{inputenc} % allow utf-8 input
\usepackage[T1]{fontenc}    % use 8-bit T1 fonts
\usepackage{url}            % simple URL typesetting
\usepackage{booktabs}       % professional-quality tables
\usepackage{multirow}       % multi-row cells in tables (\multirow)

\usepackage{amsmath}
\usepackage{amssymb} % loads amsfonts

\usepackage{microtype}      % microtypography
\usepackage{xcolor}         % colors
\usepackage{graphicx}       % images
\usepackage{colortbl}
\definecolor{citecolor}{RGB}{34,139,34}
\usepackage[pagebackref=true,breaklinks=true,letterpaper=true,colorlinks,citecolor=citecolor,bookmarks=false]{hyperref}

\newlength\savewidth\newcommand\shline{\noalign{\global\savewidth\arrayrulewidth
  \global\arrayrulewidth 1pt}\hline\noalign{\global\arrayrulewidth\savewidth}}
\newcommand{\tablestyle}[2]{\setlength{\tabcolsep}{#1}\renewcommand{\arraystretch}{#2}\centering\footnotesize}
\renewcommand{\paragraph}[1]{\vspace{1.25mm}\noindent\textbf{#1}}

\newcommand{\emojiimg}[1]{\raisebox{-0.15em}{\includegraphics[height=1.2em]{#1}}}

\title{Transforming Image Editors into Video Editors}

\author{%
Feng Wang$^{1,2}$\thanks{Work done when interning at ByteDance Seed.} ,
Zijie Li$^{1}$,
Ceyuan Yang$^{1}$,
Alan Yuille$^{2}$,
Peng Wang$^{1}$\\
$^{1}$ByteDance Seed \
$^{2}$Johns Hopkins University
}

\begin{document}

\maketitle

\begin{abstract}
Recent image editing systems have achieved impressive semantic understanding, visual fidelity, and instruction-following ability, while video editing remains substantially more difficult and costly. In this paper, we present a simple alternative to end-to-end video editing: instead of training a monolithic video editor, we transform a strong image editor into a video editor through anchor-based generation. Our key insight is that video editing can be decomposed into two subproblems: editing a sparse set of keyframes and propagating those edits across time. Based on this observation, we propose Anchor-based Video Editing (AVE), a two-stage framework in which a powerful image editor first performs composed editing on selected keyframes, and a motion-guided image-to-video diffusion model then generates the final video by treating the edited keyframes as fixed anchors. This design directly inherits the strengths of modern image editors while avoiding expensive end-to-end video editing training. Experiments on IVEBench and VIE-Bench show that AVE achieves strong performance in instruction following, temporal consistency, and content fidelity. Further ablations reveal that final video editing quality is strongly correlated with the quality of the image editor, suggesting that future progress in video editing may come from stronger image editing foundations and lightweight transfer to video. Code is available at \url{https://github.com/wangf3014/AVE}.
\end{abstract}

\section{Introduction}

Recent advances in image editing have pushed the field to a remarkable level of maturity. Modern image editing systems now exhibit strong semantic understanding, high visual fidelity, and precise instruction-following ability, often reaching a quality level suitable for real-world applications~\citep{gpt5,gemini,bagel,qwen_image_edit,instructpix2pix,flux1_kontext,seedream4,step1x_edit,omnigen}. In contrast, video editing still lags noticeably behind. Compared with image editing, video editing models must handle much longer spatiotemporal contexts, which makes them significantly more expensive to train. Meanwhile, collecting high-quality video editing data is also considerably harder, due to the need for temporal consistency, motion diversity, and reliable instruction alignment across frames. As a result, while image editing systems have rapidly advanced to a high level of quality and controllability, mainstream video editing models still remain clearly behind in visual quality, semantic consistency, and instruction-following performance~\citep{kling-omni,vace,ditto,insv2v,insvie,kiwi-edit,insedit,sama,runway,pika,minimax,diffueraser,omnivideo,lucyedit,univideo,instructx,icve,openve,vino,reco}.

To narrow this gap, this paper studies how to transfer the editing capability of image systems to video models at minimal cost. Through a series of studies, we surprisingly find that this transfer is much simpler than it appears to be. In particular, although video editing is a 3D video-to-video task, its primary challenge remains the underlying 2D image-to-image editing problem. By contrast, handling temporal information, such as preserving or modifying motion, can be achieved with relatively little additional training cost. Motivated by this observation, we introduce \underline{\textbf{A}}nchor-based \underline{\textbf{V}}ideo \underline{\textbf{E}}diting (\textbf{AVE}), a new instruction-guided video editing paradigm which explicitly decomposes the task into two stages: \textit{Composed Image Editing} and \textit{Keyframe Interpolation}.

Specifically, given an input video, we first extract a sparse set of keyframes and jointly edit them with a powerful image editing model. By performing composed image editing over multiple keyframes, we preserve scene-level and semantic consistency across the edited frames while fully benefiting from the strong instruction-following and visual generation capabilities of modern image editors. We then use an image-to-video diffusion model (\emph{e.g.,} Wan2.2-I2V~\citep{wan}) to generate the target video, where the edited keyframes serve as fixed anchors and the generation is constrained by the motion representation of the input video. In this way, the model is required to preserve the edited content at key moments while propagating the edits smoothly across intermediate frames. This decomposition provides a simple and scalable path for transforming image editors into video editors, allowing video editing systems to directly inherit the rapid progress of image editing models without requiring expensive end-to-end video editing training.

\begin{table}[t]
    \centering
    \tablestyle{5pt}{1.1}
    \vspace{-1cm}
    \begin{tabular}{l|cccc}
    Paradigm & Consistency & Frame Quality & Instruction Following & Training Cost \\\shline
    End-to-end video editing
    & Good \emojiimg{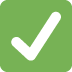}
    & Good \emojiimg{figs/check.png}
    & Depends on training \emojiimg{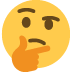}
    & Expensive \emojiimg{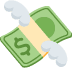} \\
    Per-frame editing
    & Bad \emojiimg{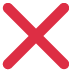}
    & Excellent \emojiimg{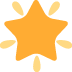}
    & Bad \emojiimg{figs/cross.png}
    & Cheap \emojiimg{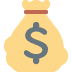} \\\rowcolor{cyan!10}
    Two-stage editing of AVE (ours)
    & Good \emojiimg{figs/check.png}
    & Excellent \emojiimg{figs/star.png}
    & Great and robust \emojiimg{figs/star.png}
    & Normal \emojiimg{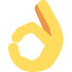} \\
    \end{tabular}
    \caption{\textbf{\textit{Intuitive comparison of three video editing paradigms.}} Our model attains favorable video consistency, frame-wise quality and instruction-following abilities with a normal cost.}
    \label{tab:paradigm_compare}
    \vspace{-0.5cm}
\end{table}

Compared with the common end-to-end video editing frameworks (\emph{e.g.}, VACE~\citep{vace}) and the most naive per-frame editing strategy, the main advantages of AVE can be summarized as follows. First, the image editing model in Stage-1 is fully decoupled from Stage-2, which allows AVE to directly leverage any mature image editor to achieve the best possible frame-wise editing quality. As a result, AVE can seamlessly inherit the rapid progress of image editing systems, including stronger semantic understanding, better visual fidelity, and more accurate instruction following, without any modification to the image editor itself. Second, unlike naive per-frame editing, which often suffers from severe flickering and temporal inconsistency, AVE performs interpolation between edited keyframes using a video diffusion model, enabling smooth propagation of edits across intermediate frames while maintaining strong temporal consistency. Third, AVE is highly cost-efficient to train. In contrast to end-to-end video editing frameworks that require substantial training resources, AVE only needs to train a lightweight motion encoder, while keeping all other components fixed and initialized from pretrained models. In our study, we find that a small motion encoder and a relatively short context length are already sufficient to capture the motion information needed for high-quality video editing. The comparison is summarized in Table~\ref{tab:paradigm_compare}.

Extensive experiments validate the effectiveness of AVE from multiple perspectives. On IVEBench~\citep{ivebench}, AVE achieves the best performance on both the short and long subsets across all reported metrics, showing that the proposed decomposition leads to a strong and well-balanced video editing capability rather than improving only a single aspect of the task. In particular, AVE is especially strong in motion-related metrics, which supports our design choice of using a motion-guided video interpolation stage to propagate edits smoothly while preserving the original temporal dynamics. AVE also obtains clear gains in semantic consistency and content fidelity, suggesting that high-quality edited anchors provide a reliable foundation for long-range video generation. On VIE-Bench~\citep{instructx}, AVE generalizes well across diverse editing types, including add, swap, remove, and style transfer, and remains competitive even against strong closed-source systems. In addition, our ablations reveal two important findings: first, the final video editing quality is strongly correlated with the quality of the Stage-1 image editor, confirming that AVE effectively transfers image editing ability into the video domain; second, AVE is robust to different keyframe selection strategies, with a moderate keyframe budget already sufficient to achieve strong performance. These results consistently support our central claim that, once high-quality edited anchors are available, the remaining image-to-video transfer is much simpler than end-to-end video editing would suggest. We hope this study provides useful insights for future video editing systems.

\section{Related Work}

\paragraph{Instruction-based image editing.}
Instruction-based image editing has advanced rapidly, evolving from early diffusion-based editors such as InstructPix2Pix~\citep{instructpix2pix} and InstructEdit~\citep{instructedit} to more capable multimodal systems such as SmartEdit~\citep{smartedit}, OmniGen~\citep{omnigen}, Step1X-Edit~\citep{step1x_edit}, and BAGEL~\citep{bagel}. Recent open and commercial image systems, including Qwen-Image-Edit~\citep{qwen_image_edit}, FLUX.1 Kontext~\citep{flux1_kontext}, and Nano Banana 2~\citep{nano_banana_2}, further improve semantic understanding, visual fidelity, and instruction following. These advances suggest that high-quality editing is increasingly solved in the image domain. Our work is complementary to this line of research: instead of proposing another image editor, we study how to transfer the editing capability of mature image systems into video models with minimal additional cost.

\paragraph{Instruction-guided video editing.}
Text-guided video editing is substantially more challenging because it must preserve temporal consistency while executing semantic edits. Early methods mainly adapt pretrained image diffusion models through inversion, one-shot tuning, or feature propagation, such as Tune-A-Video~\citep{tuneavideo}, FateZero~\citep{fatezero}, TokenFlow~\citep{tokenflow}, and InstructVideo~\citep{instructvideo}. More recent feed-forward approaches rely on synthetic datasets and end-to-end training, including InsV2V~\citep{insv2v}, InsViE~\citep{insvie}, Ditto~\citep{ditto}, VACE~\citep{vace}, Kiwi-Edit~\citep{kiwi-edit}, InsEdit~\citep{insedit}, and SAMA~\citep{sama}. Concurrent efforts also explore unified or in-context formulations and larger-scale supervision, such as ICVE~\citep{icve}, ReCo~\citep{reco}, and OpenVE-3M~\citep{openve}. While these methods have significantly improved the field, they typically require substantial video-specific training data and computation, and their editing quality still trails the best image editors. In contrast, our method does not treat video editing as a monolithic end-to-end problem. Instead, it explicitly decomposes the task into composed keyframe editing and temporal interpolation, so that the hardest part of the problem can be delegated to a strong image editor.

\paragraph{Image-to-video generation and motion modeling.}
A related line of work studies image-to-video generation and controllable motion synthesis. Representative approaches include AnimateDiff~\citep{animatediff}, VideoCrafter2~\citep{videocrafter2}, VideoComposer~\citep{videocomposer}, and methods that explicitly decouple content and motion~\citep{shen2024decouple}. These models show that motion can often be modeled separately from appearance, and that pretrained image generators can be extended to videos with relatively lightweight temporal modules. AVE builds on this insight but applies it to editing rather than pure generation. Instead of asking a video model to jointly solve semantic editing and temporal reasoning, we use the video model mainly to interpolate between edited keyframe anchors under the motion cues of the source video. This design allows AVE to inherit the rapid progress of image editing models while keeping video-specific training lightweight.

\section{Method}

\subsection{Overview}

\begin{figure}[t]
  \centering
  \includegraphics[width=\linewidth]{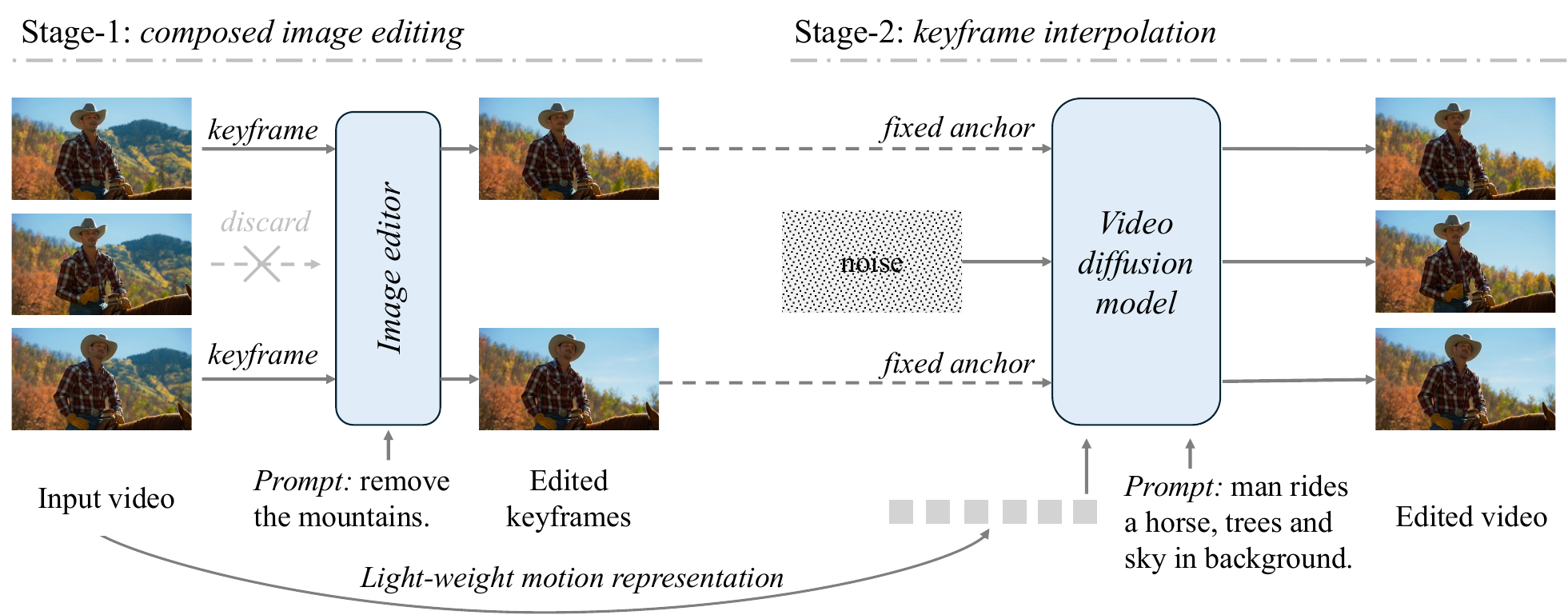}
  \caption{\textbf{\textit{Overview of our two-stage pipeline.}} In Stage-1, we edit a sparse set of keyframes using either open-source or API-based image editors. In Stage-2, an image-to-video diffusion model interpolates between the edited anchors while preserving the source motion via a lightweight motion encoder and anchor clamping.}
  \label{fig:pipeline}
\end{figure}

As illustrated in Figure~\ref{fig:pipeline}, given an input video $V=\{x_t\}_{t=1}^{T}$ and a text editing instruction $y$, our goal is to generate an edited video $\hat{V}=\{\hat{x}_t\}_{t=1}^{T}$ that faithfully follows the instruction while preserving the source motion and temporal coherence. 
Our key observation is that video editing can be decomposed into two largely independent subproblems: \emph{(i)} high-quality semantic editing on a sparse set of keyframes, and \emph{(ii)} temporally coherent interpolation between the edited keyframes. Let $\mathcal{A}=\{a_1,\dots,a_K\}$ denote the indices of $K$ selected keyframes from the source video.
AVE first edits the keyframe set $\mathbf{x}_{\mathcal{A}}=\{x_{a}\}_{a\in\mathcal{A}}$ with an image editing model $\hat{\mathbf{x}}_{\mathcal{A}} = \mathcal{E}(\mathbf{x}_{\mathcal{A}}, y)$, and then uses a video diffusion model to generate the final video conditioned on the edited keyframes as fixed anchors and a lightweight motion representation extracted from the source video $\hat{V} = \mathcal{G}(\hat{\mathbf{x}}_{\mathcal{A}}, \mathbf{m}(V))$. A key design of AVE is that Stage-1 and Stage-2 are fully decoupled: the only interface between them is the edited keyframes. 
This allows us to improve the frame-wise editing quality by simply replacing the image editor, without retraining the video generator.

\subsection{Stage-1: Composed Image Editing}

\paragraph{Training a multi-image editor from video data.}
For open-source instantiations, we train a composed image editor by adapting BAGEL~\citep{bagel} with fine-tuning.
Starting from a private video editing dataset of triplets $(V, y, V^{\star})$, where $V^{\star}$ is the edited target video, we construct multi-image editing training tuples by sampling aligned keyframes from the source and target videos $(\mathbf{x}_{\mathcal{A}}, y, \mathbf{x}^{\star}_{\mathcal{A}})$,
where $\mathbf{x}^{\star}_{\mathcal{A}}=\{x^{\star}_{a}\}_{a\in\mathcal{A}}$ denotes the target keyframes corresponding to the same timestamps.
In total, we build approximately $1.5$M multi-image editing examples from private video editing data.

Let $\mathcal{V}_{\text{img}}$ be the image VAE used by the editor, and let $\mathbf{z}_{\mathcal{A}} = \mathcal{V}_{\text{img}}(\mathbf{x}_{\mathcal{A}}),\
\mathbf{z}^{\star}_{\mathcal{A}} = \mathcal{V}_{\text{img}}(\mathbf{x}^{\star}_{\mathcal{A}})$
be the source and target latent codes.
We fine-tune BAGEL with the standard diffusion denoising objective:
\begin{equation}
\mathcal{L}_{\text{img}}
=
\mathbb{E}_{\mathbf{z}^{\star}_{\mathcal{A}},\, \epsilon,\, \tau}
\left[
\left\|
\epsilon -
\epsilon_{\phi}\!\left(
\alpha_{\tau}\mathbf{z}^{\star}_{\mathcal{A}} + \sigma_{\tau}\epsilon,\;
\mathbf{z}_{\mathcal{A}},\;
y,\;
\tau
\right)
\right\|_2^2
\right],
\end{equation}
where $\epsilon_{\phi}$ is the denoiser of the fine-tuned BAGEL, and $(\alpha_{\tau}, \sigma_{\tau})$ are the diffusion coefficients at timestep $\tau$.
By training on \emph{multiple} aligned frames jointly, the editor learns to perform consistent edits across keyframes, which is critical for downstream video generation.

\paragraph{Closed-source alternative.}
While the above open-source instantiation is useful for controlled experiments, our framework is not tied to any particular image editor.
In practice, unless otherwise specified, we instantiate Stage-1 with the latest Nano Banana API~\citep{nano_banana_2} to maximize editing quality.
For strong closed-source models, we find that the most effective strategy is to edit all selected keyframes within a single multi-turn conversation, which naturally preserves subject identity, style, and scene-level consistency across frames.
In the ablation study, we compare different API-based editors as well as open-source image editors within the same Stage-2 pipeline.

\subsection{Stage-2: Keyframe Interpolation}

Stage-2 takes the edited keyframes from Stage-1 and generates a temporally coherent target video using a pretrained image-to-video model.
We use Wan2.2-I2V-A14B~\citep{wan} as the backbone and keep all its pretrained components frozen, except for a lightweight motion encoder and its associated control branch.

\begin{figure}[t]
  \centering
  \includegraphics[width=0.6\linewidth]{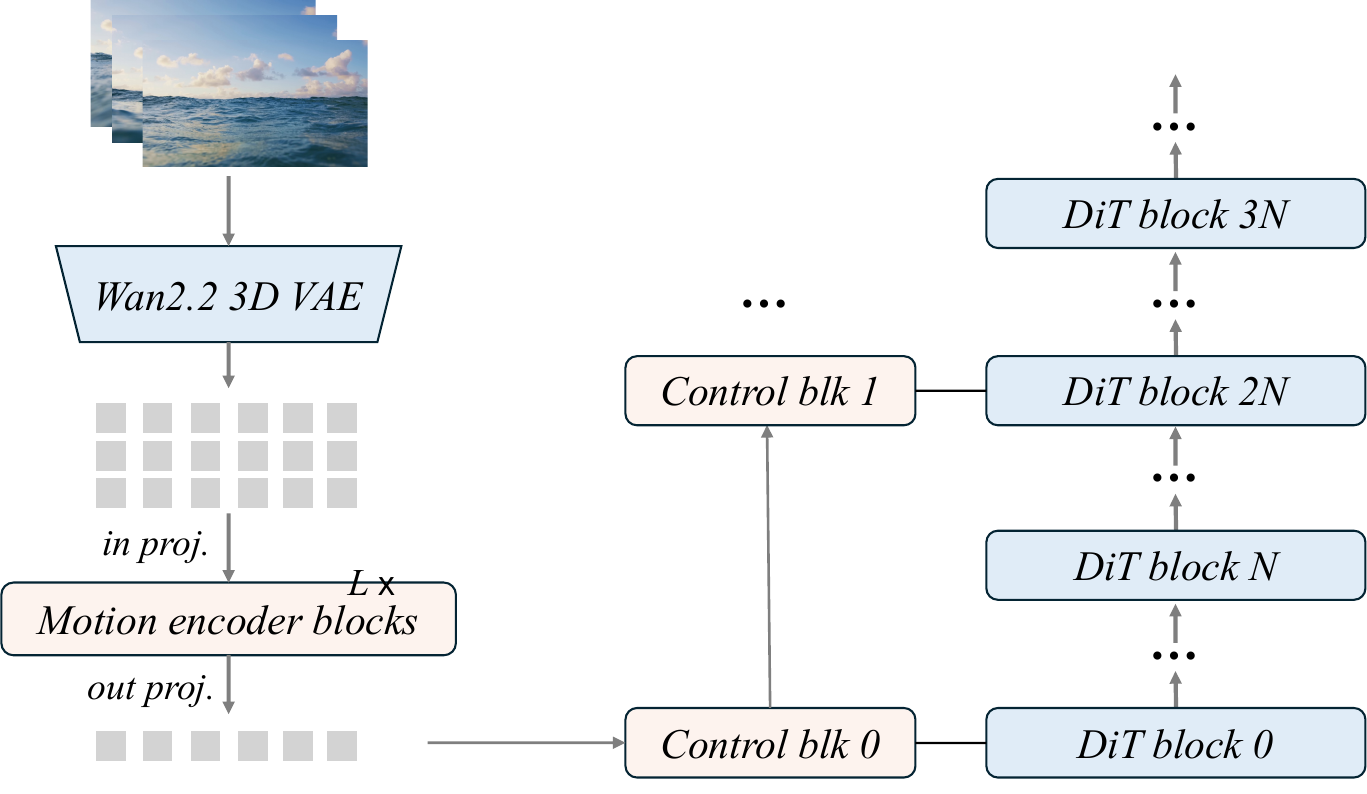}
  \caption{Architecture of the lightweight motion encoder used in Stage-2. The encoder projects video latents into a token sequence and applies factorized spatial-then-temporal self-attention to extract motion features that condition the frozen image-to-video backbone.}
  \label{fig:motion}
\end{figure}

\paragraph{Lightweight motion encoder.} Let $\mathcal{V}_{3\text{D}}^{\text{enc}}$ and $\mathcal{V}_{3\text{D}}^{\text{dec}}$ denote the 3D VAE encoder and decoder of Wan2.2-I2V-A14B.
The source video is encoded into latent space as $Z = \mathcal{V}_{3\text{D}}^{\text{enc}}(V) \in \mathbb{R}^{T \times H \times W \times C}.$
We project latent tokens into the motion embedding space: $U^{(0)}_{t,h,w} = W_{\text{in}} Z_{t,h,w} + e_t^{\text{temp}} + e_{h,w}^{\text{spatial}}$, where $W_{\text{in}} \in \mathbb{R}^{C \times d}$ is a learnable linear projection, and $e_t^{\text{temp}}$ and $e_{h,w}^{\text{spatial}}$ denote temporal and spatial positional embeddings, respectively. 
The resulting tensor $U^{(0)} \in \mathbb{R}^{T \times H \times W \times d}$ is then processed by a 6-layer factorized spatiotemporal Transformer. Each layer consists of a spatial self-attention block followed by a temporal self-attention block and an MLP. We inject the final motion representation into the frozen Wan2.2 DiT through a ControlNet-style residual branch~\citep{controlnet}, allowing the generator to preserve source motion while leaving appearance and editing semantics to be determined by the edited anchors. The overall framework of the motion encoder is shown in Figure~\ref{fig:motion}. More details about the architecture and training approach can be found in Appendix.

\paragraph{Anchor-preserving diffusion.}
Let $D_{\psi}$ denote the pretrained Wan2.2 DiT.
During generation, the edited keyframes are treated as \emph{fixed anchors}: their latent features are clamped throughout the entire reverse diffusion process and are never denoised. Formally, we define a binary anchor mask
\begin{equation}
M_{\mathcal{A}} \in \{0,1\}^{T \times H \times W \times C},
\end{equation}
where $M_{\mathcal{A}}(t)=1$ if frame $t \in \mathcal{A}$ and $0$ otherwise. We encode the anchor frames by Wan2.2's 3D VAE encoder:
\begin{equation}
    \hat{Z}_{\mathcal{A}} = \mathcal{V}_{3\text{D}}^{\text{enc}}(\hat{\mathbf{x}}_{\mathcal{A}}).
\end{equation}
At the initial diffusion step $S$, the latent video is initialized as
\begin{equation}
X_S
=
M_{\mathcal{A}} \odot \hat{Z}_{\mathcal{A}}
+
(1-M_{\mathcal{A}})\odot \epsilon,
\qquad
\epsilon \sim \mathcal{N}(0, I),
\end{equation}
that is, anchor frames are directly set to the edited latent anchors, while all non-anchor frames start from random noise. At each reverse diffusion step $s$, the frozen DiT predicts the noise residual conditioned on the current latent video and the motion control:
\begin{equation}
\hat{\epsilon}_s = D_{\psi}(X_s, h_{\text{motion}}).
\end{equation}
A standard diffusion update produces an intermediate estimate
\begin{equation}
\tilde{X}_{s-1} = \mathrm{Step}(X_s, \hat{\epsilon}_s),
\end{equation}
after which the anchor frames are re-imposed:
\begin{equation}
X_{s-1}
=
M_{\mathcal{A}} \odot \hat{Z}_{\mathcal{A}}
+
(1-M_{\mathcal{A}})\odot \tilde{X}_{s-1}.
\end{equation}
Therefore, anchor features remain unchanged throughout denoising, while the model fills in all intermediate frames conditioned on both the fixed visual anchors and the source motion representation. After the final denoising step, the edited video is decoded back to pixel space by the 3D VAE decoder:
\begin{equation}
\hat{V} = \mathcal{V}_{3\text{D}}^{\text{dec}}(X_0).
\end{equation}

\section{Experiments}
\subsection{Experimental Setup}

We use Wan2.2-I2V-A14B~\citep{wan} as the video generation backbone. Unless otherwise specified, all pretrained components in Wan, including the 3D VAE encoder/decoder and the DiT generator, are kept frozen during training. The total number of learnable parameters is 710M, which is about 5\% of the diffusion model's parameter count. In our experiments, we find that only lightweight training on the motion-related components is sufficient to achieve strong motion preservation and frame-to-video generation performance. This is mainly because we already effectively leverage the motion modeling and understanding capability of Wan2.2's pretrained 3D VAE and diffusion transformer. In practice, we train the motion encoder and ControlNet on only 500K video editing samples, and do not observe further improvements when scaling up either the data size or the training duration. We train our model on 64 A100 GPUs. We optimize the learnable parameters using AdamW with a base learning rate of $1\times10^{-4}$ and weight decay $0.05$.
The optimizer uses $\beta_1=0.9$ and $\beta_2=0.999$.
We apply a cosine learning rate schedule with linear warmup at the beginning of training.
All experiments are conducted in mixed precision. We evaluate our method on two public benchmarks, IVEBench~\citep{ivebench} and VIEBench~\citep{instructx}, which assess instruction-guided video editing from different perspectives.
Unless otherwise specified, we follow the standard evaluation protocols of each benchmark and report their official metrics.

\begin{table}[t]
    \centering
    \tablestyle{6pt}{1.1}
    \begin{tabular}{l|cccc|ccc}
    Model & Total & Quality & Ins. Comp. & Fidelity & Motion & Sem. Consist. & Content \\\shline
    Ditto~\citep{ditto} & 66.75 & 78.12 & 49.08 & 73.04 & 78.99 & 24.92 & 3.64 \\
    InsV2V~\citep{insv2v} & 66.68 & 79.58 & 38.61 & 81.85 & 85.55 & 24.11 & 4.05 \\
    Lucy-Edit-Dev~\citep{lucyedit} & 63.53 & 82.09 & 33.94 & 74.55 & 67.79 & 23.83 & 3.83 \\
    VACE~\citep{vace} & 62.61 & 79.83 & 25.42 & \bf 82.58 & 88.59 & 23.80 & 4.03 \\
    ICVE~\citep{icve} & 60.33 & 71.25 & 45.35 & 64.37 & 45.72 & 22.86 & 3.55 \\
    Omni-Video~\citep{omnivideo} & 58.65 & 78.20 & 43.64 & 54.11 & 50.72 & 21.98 & 2.85 \\
    AnyV2V~\citep{anyv2v} & 57.67 & 72.71 & 41.72 & 58.59 & 81.59 & 21.59 & 2.75 \\
    StableV2V~\citep{stablev2v} & 50.89 & 69.17 & 42.67 & 40.84 & 75.13 & 19.78 & 1.79 \\
    \rowcolor{cyan!10}
    AVE (ours) & \bf 70.58 & \bf 82.10 & \bf 49.65 & 80.01 & \bf 89.26 & \bf 26.12 & \bf 4.10 \\
    \end{tabular}
    \caption{\textit{\textbf{IVEBench short subset results.}} Here we report the overall performance of total score, video quality, instruction compliance, video fidelity, and detailed metrics of motion smoothness, semantic consistency, and content fidelity and leave the remaining scores of IVEBench in Appendix. Our AVE model consistently obtains the best results among the evaluation metrics.}
    \label{tab:ivebench_short_main}
\end{table}

\begin{table}[t]
    \centering
    \tablestyle{6pt}{1.1}
    \begin{tabular}{l|cccc|ccc}
    Model & Total & Quality & Ins. Comp. & Fidelity & Motion & Sem. Consist. & Content \\\shline
    Ditto~\citep{ditto} & 65.93 & 77.95 & 47.85 & 71.98 & 72.82 & 23.42 & 3.69 \\
    InsV2V~\citep{insv2v} & 65.72 & 80.24 & 37.41 & 79.50 & 67.88 & 24.06 & 4.13 \\
    Lucy-Edit-Dev~\citep{lucyedit} & 64.89 & 82.10 & 31.51 & \bf 81.05 & 73.46 & 23.92 & 4.13 \\
    VACE~\citep{vace} & 61.61 & 80.12 & 26.73 & 77.98 & 88.40 & 23.56 & 3.74 \\
    ICVE~\citep{icve} & 58.77 & 71.96 & 40.38 & 63.98 & 48.40 & 22.60 & 3.48 \\
    Omni-Video~\citep{omnivideo} & 57.09 & 77.84 & 42.44 & 51.00 & 55.11 & 21.90 & 2.59 \\
    AnyV2V~\citep{anyv2v} & 55.05 & 72.40 & 35.55 & 57.20 & 82.47 & 21.55 & 2.65 \\
    StableV2V~\citep{stablev2v} & 50.93 & 69.37 & 42.09 & 41.33 & 77.33 & 20.41 & 1.79 \\
    \rowcolor{cyan!10}
    AVE (ours) & \bf 68.75 & \bf 82.15 & \bf 48.31 & 75.79 & \bf 88.95 & \bf 24.68 & \bf 4.16 \\
    \end{tabular}
    \caption{\textit{\textbf{IVEBench long subset results.}} Similar to the short subset, our model attains the best results across the reported metrics. AVE has good robustness to long-video editing.}
    \label{tab:ivebench_long_main}
\end{table}

\subsection{Main Results}

Tables~\ref{tab:ivebench_short_main} and~\ref{tab:ivebench_long_main} report the main results on the short and long subsets of IVEBench, respectively. 
AVE achieves the best performance on most reported metrics on both subsets, showing that our two-stage design leads to a strong and well-balanced video editing capability rather than improving only a single aspect of the task. 
On the short subset, AVE improves the total score from the previous best 66.75 to 70.58, while also establishing new best results in video quality (82.10), instruction compliance (49.65), and semantic consistency (26.12). 
The same trend holds on the long subset, where AVE reaches a total score of 68.75, clearly outperforming the previous best result. 
These results suggest that AVE does not merely optimize perceptual quality or instruction following; instead, it achieves a favorable balance among edit faithfulness, temporal coherence, and source-video preservation.

A key strength of AVE lies in motion-related quality. 
On IVEBench, AVE obtains the best motion score on both the short and long subsets, reaching 89.26 and 88.95, respectively. 
We attribute this advantage to the design of Stage-2. 
Instead of asking the video model to jointly solve semantic editing and temporal reasoning from scratch, AVE treats video generation as an anchor-constrained interpolation problem. 
The edited keyframes explicitly define the desired content at several time steps, while the Wan-based video generator fills in the intermediate frames under the guidance of a motion representation extracted from the source video. 
This division of labor makes the temporal generation process much more stable, and naturally reduces flickering, drift, and motion inconsistency that commonly appear in naive per-frame editing and end-to-end video editing systems. 
In other words, AVE preserves motion well precisely because the video model is used for what it is best at: temporally coherent interpolation rather than full semantic editing.

Table~\ref{tab:viebench_main} further shows that AVE generalizes well to VIE-Bench, which evaluates diverse editing types including add, swap/change, remove, and style/tone transfer. 
AVE is particularly strong on \emph{swap/change} and \emph{remove}, where it achieves the best average scores of 9.448 and 9.352, respectively. 
We believe these categories benefit the most from anchor-based generation, because the edited keyframes explicitly specify which object or region should be replaced or removed, and the video model only needs to propagate the edited result consistently through time. 
AVE also achieves the best averaged instruction-following score (9.539) and preservation score (9.425) across categories, indicating that it not only executes the edit faithfully, but also preserves the remaining content better than prior methods. 
This is exactly the behavior we would expect from our decomposition: Stage-1 handles the difficult semantic edit with a strong image editor, while Stage-2 focuses on preserving motion and completing the video around those edited anchors.

It is also worth noting that AVE remains competitive even against strong closed-source commercial systems. 
Although the best closed-source models still have a slight edge on a few appearance-centric categories such as \emph{add} and \emph{style/tone}, AVE already surpasses them on several important dimensions, including swap/change, remove, and the averaged instruction-following and preservation metrics. 
At the same time, AVE achieves a competitive overall quality score of 8.630, which is on par with or slightly better than the strongest open-source baselines. 
Taken together, these results show that AVE is not simply a lightweight approximation to end-to-end video editing. 
Rather, it provides a more effective formulation of the problem, enabling a video editor to directly inherit the strengths of modern image editing systems while retaining strong temporal coherence through a lightweight video generation stage.

\begin{table}[t]
    \centering
    \tablestyle{5pt}{1.1}
    \begin{tabular}{l|cccc|ccc}
    Method & Add & Swap / Change & Remove & Style / Tone & Ins. Follow & Pres. & Quality \\\shline
    \rowcolor{gray!10}
    Kling1.6~\citep{kling-omni} & 6.602 & 8.800 & 8.253 & - & 7.813 & 8.697 & 7.143 \\
    \rowcolor{gray!10}
    Kling-Omni~\citep{kling-omni} & \textbf{9.181} & 9.194 & 9.133 & \textbf{9.341} & 9.518 & 9.346 & \textbf{8.751} \\
    \rowcolor{gray!10}
    Runway~\citep{runway} & 8.447 & 9.161 & 8.504 & 9.133 & 9.109 & 8.972 & 8.354 \\
    \rowcolor{gray!10}
    Pika~\citep{pika} & - & 7.408 & - & - & 7.542 & 7.847 & 6.837 \\
    \rowcolor{gray!10}
    MiniMax~\citep{minimax} & - & - & 6.839 & - & 6.963 & 7.518 & 6.037 \\
    DiffuEraser~\citep{diffueraser} & - & - & 6.243 & - & 6.346 & 6.807 & 5.576 \\
    InsV2V~\citep{insv2v} & 4.281 & 5.567 & 2.098 & 7.452 & 4.475 & 6.043 & 4.033 \\
    VACE~\citep{vace} & 4.854 & 6.640 & 2.682 & - & 3.974 & 6.042 & 4.162 \\
    Omni-Video~\citep{omnivideo} & 6.242 & 4.748 & 5.593 & 5.366 & 5.481 & 5.404 & 5.429 \\
    UniVideo~\citep{univideo} & 8.656 & 8.683 & 8.233 & 8.711 & 8.708 & 8.963 & 8.042 \\
    InstructX~\citep{instructx} & 8.349 & 9.072 & 8.322 & 9.196 & 9.059 & 8.905 & 8.241 \\
    SAMA~\citep{sama} & 8.711 & 9.340 & 9.144 & 9.259 & 9.344 & 9.370 & 8.626 \\
    \rowcolor{cyan!10}
    AVE (ours) & 9.054 & \bf 9.448 & \bf 9.352 & 9.203 & \bf 9.539 & \bf 9.425 & 8.630 \\
    \end{tabular}
    \caption{\textit{\textbf{Comparison results on VIE-Bench.}} 
    We report the average score of each editing category (add, swap, remove, and style) and each evaluation metric (instruction following, motion preservation, and video quality). Results of closed-source commercial models are marked in \textcolor{gray}{gray}.}
    \label{tab:viebench_main}
\end{table}

\begin{figure}[t]
  \centering
  \includegraphics[width=\textwidth]{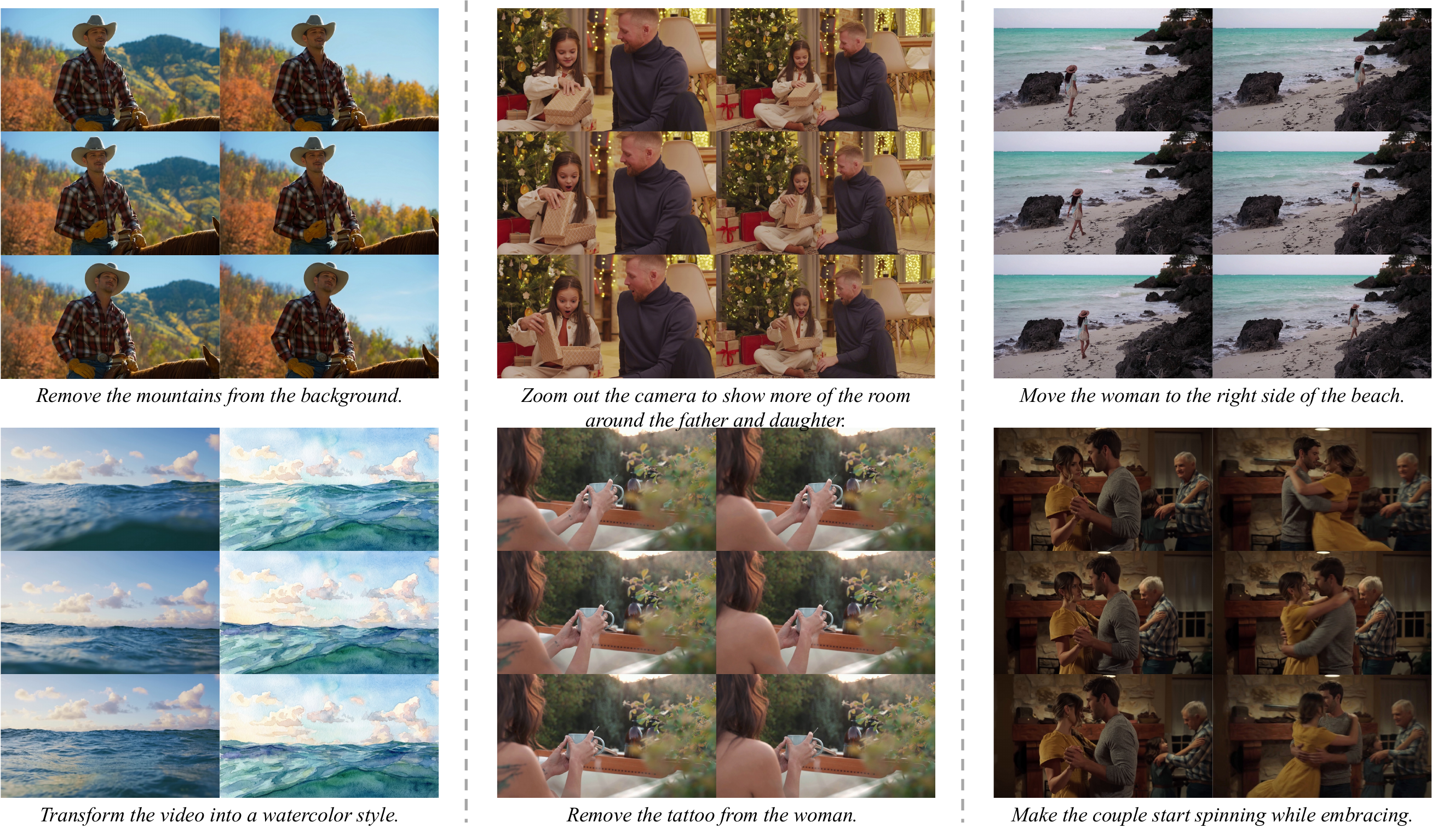}
  \caption{\textbf{\textit{Instruction-based video editing results produced by AVE.}} Guided by keyframe-level instructions and image-to-video extension, our method delivers high-fidelity visual details, strong temporal coherence, and robust performance across diverse editing tasks.}
  \label{fig:teaser}
\end{figure}

\paragraph{Qualitative results.} We provide qualitative examples in Fig.~\ref{fig:teaser} to further illustrate the versatility of AVE across a wide range of editing scenarios. 
Our method performs well on diverse instruction types, including background removal, viewpoint change, object relocation, style transfer, local attribute editing, and explicit motion modification. 
A notable advantage of AVE is that, when the edit mainly changes appearance or scene content, the original motion and temporal structure are preserved very well. 
For example, in the first example, where the mountains are removed from the background, the rider’s pose and motion remain nearly unchanged across frames, while the new background stays consistent and coherent over time. 
In the zoom-out example, AVE expands the room around the father and daughter while maintaining stable identity, posture, and scene layout. 
The beach example further shows that AVE can handle object relocation reliably: the woman is moved to the right side of the frame, while the coastline and ocean remain temporally coherent.

The second row demonstrates that AVE is also effective on more challenging edits. 
In the watercolor example, the style is transformed consistently across frames while preserving the wave structure and overall scene composition, showing that AVE can handle global appearance changes rather than only local edits. 
In the tattoo removal example, the model performs a subtle local modification while keeping the surrounding body shape and background stable, indicating strong fine-grained control. 
Most importantly, the final example shows that AVE can support substantial motion changes: the couple is edited to start spinning while embracing, yet the generated motion remains smooth and stable instead of becoming abrupt or jittery. 
Together, these examples highlight the core strength of AVE: strong semantic edits are established at the keyframe level, and the motion-guided interpolation stage then propagates them through time in a controlled and coherent manner.

\subsection{Ablation Study}

Unless otherwise specified, all ablations are conducted on the long subset of IVEBench. Each experiment is done by three runs and the standard deviation is less than 0.1\%. We focus on two questions: (1) how much the final video editing performance depends on the image editor used in Stage-1, and (2) how sensitive AVE is to the keyframe selection strategy.

\paragraph{Effect of the Stage-1 image editor.}
Table~\ref{tab:abl_editor} studies the impact of replacing the image editor in Stage-1 while keeping Stage-2 unchanged.
By default, AVE uses Nano Banana 2, which yields the best overall results.
Replacing it with other strong closed-source editors such as GPT-5.1 or Seedream 4.0 only causes a mild drop, while switching to our finetuned BAGEL leads to a further decrease. Using the original untuned BAGEL causes a much larger degradation. This trend provides a particularly important insight: the editing ability of AVE is strongly correlated with the quality of the image editor.
Notably, the largest drops appear in instruction compliance, semantic consistency, and content fidelity, while motion remains relatively stable.
This is consistent with our design: motion is mainly handled by Stage-2, which is fixed across all variants, whereas the semantic correctness of the edit is largely determined by the anchor frames produced by Stage-1.
In other words, better image editing almost directly translates into better video editing.
This strongly supports our main claim that the more difficult part of video editing is still the underlying image editing problem, while the image-to-video transfer itself is comparatively simple once high-quality anchors are available.

It is also worth noting that the tuned BAGEL model significantly outperforms the original untuned BAGEL.
This suggests that multi-image consistent editing is important even before the video generation stage, and validates our choice of adapting open-source image editors with multi-frame video-derived supervision when closed-source APIs are unavailable.

\begin{table}[t]
    \centering
    \tablestyle{5pt}{1.1}
    \begin{tabular}{l|cccc|ccc}
    Stage-1 Image Editor & Total & Quality & Ins. Comp. & Fidelity & Motion & Sem. Consist. & Content \\\shline\rowcolor{cyan!10}
    Nano Banana 2 (default) & \bf 68.75 & \bf 82.15 & \bf 48.31 & \bf 75.79 & \bf 88.95 & \bf 24.68 & \bf 4.16 \\
    GPT-5.1 & 68.10 & 81.43 & 47.11 & 75.76 & 88.74 & 24.10 & 4.02 \\
    Seedream 4.0 & 67.59 & 80.36 & 46.67 & 75.74 & 88.61 & 23.83 & 3.85 \\
    BAGEL (finetuned) & 66.07 & 79.38 & 43.06 & 75.78 & 88.10 & 21.96 & 3.83 \\
    BAGEL (base) & 58.54 & 68.12 & 37.84 & 69.66 & 77.21 & 20.34 & 3.42 \\
    \end{tabular}
    \caption{\textit{\textbf{Ablation on the Stage-1 image editor on IVEBench-long.}}
    We replace only the composed image editor in Stage-1 and keep the entire Stage-2 pipeline fixed. The results show a strong correlation between image editing quality and final video editing performance.}
    \label{tab:abl_editor}
\end{table}

\paragraph{Effect of keyframe selection.}
Table~\ref{tab:abl_kf} compares different keyframe selection strategies and budgets.
Our default strategy first extracts scene-aware anchors using PySceneDetect, keeps the first frame of each scene together with the final frame of the video, and then fills the remaining budget using a largest-gap rule.
We compare this strategy against simple uniform sampling under different keyframe budgets. Two observations are clear.
First, the gap between the two strategies is relatively small across all settings, indicating that AVE is fairly robust to the exact keyframe selection rule.
This is encouraging, as it suggests that once a sparse set of reasonable anchors is provided, the Stage-2 interpolation model can reliably propagate the edits to the full video.
Still, our scene-aware strategy is consistently better than uniform sampling, especially when the budget is limited, likely because it is more likely to place anchors near semantic transitions and scene changes.

Second, the best performance is achieved when the number of keyframes is set to approximately $\lfloor T/24 \rfloor$.
Using too few keyframes leads to weaker instruction compliance and lower fidelity because the anchors become too sparse to fully specify the desired edit over long sequences.
On the other hand, using too many keyframes also hurts performance slightly.
We hypothesize that this is because the Wan2.2 3D VAE already performs a $4\times$ temporal downsampling, so overly dense anchors reduce the effective interpolation space and make temporal completion less stable.
As a result, AVE performs best when the anchors are sparse enough to allow interpolation, but dense enough to stabilize the desired edit.

\begin{table}[t]
    \centering
    \tablestyle{5pt}{1.1}
    \vspace{-0.5cm}
    \begin{tabular}{l|c|ccccc}
    Strategy & \#Keyframes & Total & Quality & Ins. Comp. & Fidelity & Motion \\\shline
    Uniform & 2 & 66.85 & 79.58 & 46.02 & 74.96 & 87.42 \\
    Uniform & $\lfloor T/48 \rfloor$ & 67.73 & 80.84 & 47.12 & 75.24 & 88.21 \\
    Uniform & $\lfloor T/24 \rfloor$ & 68.21 & 81.69 & 47.86 & 75.09 & 88.71 \\
    Uniform & $\lfloor T/12 \rfloor$ & 67.94 & 81.27 & 47.40 & 75.15 & 88.38 \\\hline
    Scene-first + max-gap (ours) & 2 & 67.28 & 79.93 & 46.48 & 75.44 & 87.67 \\
    Scene-first + max-gap (ours) & $\lfloor T/48 \rfloor$ & 68.09 & 81.16 & 47.56 & 75.55 & 88.46 \\\rowcolor{cyan!10}
    Scene-first + max-gap (ours) & $\lfloor T/24 \rfloor$ & \bf 68.75 & \bf 82.15 & \bf 48.31 & \bf 75.79 & \bf 88.95 \\
    Scene-first + max-gap (ours) & $\lfloor T/12 \rfloor$ & 68.34 & 81.61 & 47.83 & 75.59 & 88.54 \\
    \end{tabular}
    \caption{\textit{\textbf{Ablation on keyframe selection strategies and keyframe budgets on IVEBench-long.}}
    Our default strategy combines scene-aware anchors from PySceneDetect with largest-gap completion. AVE is robust to the exact keyframe selection rule, while a moderate keyframe budget around $\lfloor T/24 \rfloor$ gives the best trade-off.}
    \label{tab:abl_kf}
    \vspace{-0.5cm}
\end{table}

\section{Conclusion}
In this paper, we propose Anchor-based Video Editing (AVE), a simple two-stage framework that transforms image editors into video editors. Our key insight is that the main difficulty of video editing still lies in image-level semantic editing, while temporal propagation can be handled much more efficiently once a sparse set of high-quality edited keyframes is available. Based on this observation, AVE first performs composed image editing on selected keyframes, and then uses a motion-guided image-to-video diffusion model to interpolate the full edited video. This design directly inherits the strengths of modern image editing systems while avoiding expensive end-to-end video editing training. Experiments on IVEBench and VIE-Bench show that AVE achieves strong performance across instruction following, semantic consistency, content fidelity, and motion preservation, and our ablations further reveal that final video editing quality is strongly correlated with the quality of the image editor. We hope this work offers useful insights into building future video editing systems in a more modular, efficient, and scalable way.

{\small
\bibliographystyle{ieeenat_fullname}
\bibliography{main}

@String(CVPR  = {CVPR})

@String(ICCV  = {ICCV})

@String(ECCV  = {ECCV})

@String(NIPS  = {NeurIPS})

@String(ICLR  = {ICLR})

@String(ICML  = {ICML})

@String(AAAI  = {AAAI})

@article{wan,
  title={{Wan}: Open and Advanced Large-Scale Video Generative Models},
  author={{Wan Team} and Wang, Ang and Ai, Baole and Wen, Bin and Mao, Chaojie and Xie, Chen-Wei and Chen, Di and Yu, Feiwu and Zhao, Haiming and Yang, Jianxiao and others},
  journal={arXiv preprint arXiv:2503.20314},
  year={2025}
}

@inproceedings{videocomposer,
  title={{VideoComposer}: Compositional Video Synthesis with Motion Controllability},
  author={Wang, Xiang and Yuan, Hangjie and Zhang, Shiwei and Chen, Dayou and Wang, Jiuniu and Zhang, Yingya and Shen, Yujun and Zhao, Deli and Zhou, Jingren},
  booktitle=NIPS,
  year={2023}
}

@inproceedings{shen2024decouple,
  title={Decouple Content and Motion for Conditional Image-to-Video Generation},
  author={Shen, Cuifeng and Gan, Yulu and Chen, Chen and Zhu, Xiongwei and Cheng, Lele and Gao, Tingting and Wang, Jinzhi},
  booktitle=AAAI,
  year={2024}
}

@inproceedings{instructvideo,
  title={{InstructVideo}: Instructing Video Diffusion Models with Human Feedback},
  author={Yuan, Hangjie and Zhang, Shiwei and Wang, Xiang and Wei, Yujie and Feng, Tao and Pan, Yining and Zhang, Yingya and Liu, Ziwei and Albanie, Samuel and Ni, Dong},
  booktitle=CVPR,
  year={2024}
}

@inproceedings{animatediff,
  title={{AnimateDiff}: Animate Your Personalized Text-to-Image Diffusion Models without Specific Tuning},
  author={Guo, Yuwei and Yang, Ceyuan and Rao, Anyi and Liang, Zhengyang and Wang, Yaohui and Qiao, Yu and Agrawala, Maneesh and Lin, Dahua and Dai, Bo},
  booktitle=ICLR,
  year={2024}
}

@inproceedings{videocrafter2,
  title={{VideoCrafter2}: Overcoming Data Limitations for High-Quality Video Diffusion Models},
  author={Chen, Haoxin and Zhang, Yong and Cun, Xiaodong and Xia, Menghan and Wang, Xintao and Weng, Chao and Shan, Ying},
  booktitle=CVPR,
  year={2024}
}

@inproceedings{controlnet,
  title={Adding Conditional Control to Text-to-Image Diffusion Models},
  author={Zhang, Lvmin and Rao, Anyi and Agrawala, Maneesh},
  booktitle=ICCV,
  year={2023}
}

@article{kling-omni,
  title={{Kling-Omni} Technical Report},
  author={{Kling Team} and Chen, Jialu and Ci, Yuanzheng and Du, Xiangyu and Feng, Zipeng and Gai, Kun and Guo, Sainan and Han, Feng and He, Jingbin and He, Kang and others},
  journal={arXiv preprint arXiv:2512.16776},
  year={2025}
}

@inproceedings{vace,
  title={{VACE}: All-in-One Video Creation and Editing},
  author={Jiang, Zeyinzi and Han, Zhen and Mao, Chaojie and Zhang, Jingfeng and Pan, Yulin and Liu, Yu},
  booktitle=ICCV,
  year={2025}
}

@inproceedings{ditto,
  title={Scaling Instruction-Based Video Editing with a High-Quality Synthetic Dataset},
  author={Bai, Qingyan and Wang, Qiuyu and Ouyang, Hao and Yu, Yue and Wang, Hanlin and Wang, Wen and Cheng, Ka Leong and Ma, Shuailei and Zeng, Yanhong and Liu, Zichen and Xu, Yinghao and Shen, Yujun and Chen, Qifeng},
  booktitle=CVPR,
  year={2026}
}

@inproceedings{tokenflow,
  title={{TokenFlow}: Consistent Diffusion Features for Consistent Video Editing},
  author={Geyer, Michal and Bar-Tal, Omer and Bagon, Shai and Dekel, Tali},
  booktitle=ICLR,
  year={2024}
}

@inproceedings{insv2v,
  title={Consistent Video-to-Video Transfer Using Synthetic Dataset},
  author={Cheng, Jiaxin and Xiao, Tianjun and He, Tong},
  booktitle=ICLR,
  year={2024}
}

@inproceedings{insvie,
  title={{InsViE-1M}: Effective Instruction-Based Video Editing with Elaborate Dataset Construction},
  author={Wu, Yuhui and Chen, Liyi and Li, Ruibin and Wang, Shihao and Xie, Chenxi and Zhang, Lei},
  booktitle=ICCV,
  year={2025}
}

@article{kiwi-edit,
  title={{Kiwi-Edit}: Versatile Video Editing via Instruction and Reference Guidance},
  author={Lin, Yiqi and Liang, Guoqiang and Zeng, Ziyun and Bai, Zechen and Chen, Yanzhe and Shou, Mike Zheng},
  journal={arXiv preprint arXiv:2603.02175},
  year={2026}
}

@article{insedit,
  title={{InsEdit}: Towards Instruction-based Visual Editing via Data-Efficient Video Diffusion Models Adaptation},
  author={Rao, Zhefan and Zou, Bin and Che, Haoxuan and He, Xuanhua and Choi, Chong Hou and Li, Yanheng and Liu, Rui and Chen, Qifeng},
  journal={arXiv preprint arXiv:2604.08646},
  year={2026}
}

@article{sama,
  title={{SAMA}: Factorized Semantic Anchoring and Motion Alignment for Instruction-Guided Video Editing},
  author={Zhang, Xinyao and Dong, Wenkai and Song, Yuxin and Fang, Bo and Zhang, Qi and Wang, Jing and Chen, Fan and Zhang, Hui and Feng, Haocheng and Lu, Yu and others},
  journal={arXiv preprint arXiv:2603.19228},
  year={2026}
}

@misc{runway,
  author={{Runway}},
  title={Introducing {Runway Gen-4}},
  year={2025},
  howpublished={\url{https://runwayml.com/research/introducing-runway-gen-4}}
}

@misc{pika,
  author={{Pika}},
  title={{Pika}},
  year={2026},
  howpublished={\url{https://pika.art/}}
}

@inproceedings{minimax,
  title={{MiniMax-Remover}: Taming Bad Noise Helps Video Object Removal},
  author={Zi, Bojia and Peng, Weixuan and Qi, Xianbiao and Wang, Jianan and Zhao, Shihao and Xiao, Rong and Wong, Kam-Fai},
  booktitle=NIPS,
  year={2025}
}

@article{diffueraser,
  title={{DiffuEraser}: A Diffusion Model for Video Inpainting},
  author={Li, Xiaowen and Xue, Haolan and Ren, Peiran and Bo, Liefeng},
  journal={arXiv preprint arXiv:2501.10018},
  year={2025}
}

@article{omnivideo,
  title={{Omni-Video}: Democratizing Unified Video Understanding and Generation},
  author={Tan, Zhiyu and Yang, Hao and Qin, Luozheng and Gong, Jia and Yang, Mengping and Li, Hao},
  journal={arXiv preprint arXiv:2507.06119},
  year={2025}
}

@misc{lucyedit,
  title={{Lucy Edit}: Open-Weight Text-Guided Video Editing},
  author={{DecartAI Team}},
  year={2025},
  howpublished={\url{https://huggingface.co/decart-ai/Lucy-Edit-Dev}}
}

@inproceedings{univideo,
  title={{UniVideo}: Unified Understanding, Generation, and Editing for Videos},
  author={Wei, Cong and Liu, Quande and Ye, Zixuan and Wang, Qiulin and Wang, Xintao and Wan, Pengfei and Gai, Kun and Chen, Wenhu},
  booktitle=ICLR,
  year={2026}
}

@article{instructx,
  title={{InstructX}: Towards Unified Visual Editing with {MLLM} Guidance},
  author={Mou, Chong and Sun, Qichao and Wu, Yanze and Zhang, Pengze and Li, Xinghui and Ye, Fulong and Zhao, Songtao and He, Qian},
  journal={arXiv preprint arXiv:2510.08485},
  year={2025}
}

@article{icve,
  title={In-Context Learning with Unpaired Clips for Instruction-Based Video Editing},
  author={Liao, Xinyao and Zeng, Xianfang and Song, Ziye and Fu, Zhoujie and Yu, Gang and Lin, Guosheng},
  journal={arXiv preprint arXiv:2510.14648},
  year={2025}
}

@inproceedings{openve,
  title={{OpenVE-3M}: A Large-Scale High-Quality Dataset for Instruction-Guided Video Editing},
  author={He, Haoyang and Wang, Jie and Zhang, Jiangning and Xue, Zhucun and Bu, Xingyuan and Yang, Qiangpeng and Wen, Shilei and Xie, Lei},
  booktitle=ECCV,
  year={2026}
}

@article{vino,
  title={{VINO}: A Unified Visual Generator with Interleaved {OmniModal} Context},
  author={Chen, Junyi and He, Tong and Fu, Zhoujie and Wan, Pengfei and Gai, Kun and Ye, Weicai},
  journal={arXiv preprint arXiv:2601.02358},
  year={2026}
}

@inproceedings{reco,
  title={In-Context Generation with Regional Constraints for Instructional Video Editing},
  author={Zhang, Zhongwei and Long, Fuchen and Li, Wei and Qiu, Zhaofan and Liu, Wu and Yao, Ting and Mei, Tao},
  booktitle=ICML,
  year={2026}
}

@inproceedings{ivebench,
  title={{IVEBench}: Modern Benchmark Suite for Instruction-Guided Video Editing Assessment},
  author={Chen, Yinan and Zhang, Jiangning and Hu, Teng and Zeng, Yuxiang and Xue, Zhucun and He, Qingdong and Wang, Chengjie and Liu, Yong and Hu, Xiaobin and Yan, Shuicheng},
  booktitle=ICLR,
  year={2026}
}

@inproceedings{tuneavideo,
  title={{Tune-A-Video}: One-Shot Tuning of Image Diffusion Models for Text-to-Video Generation},
  author={Wu, Jay Zhangjie and Ge, Yixiao and Wang, Xintao and Lei, Weixian and Gu, Yuchao and Shi, Yufei and Hsu, Wynne and Shan, Ying and Qie, Xiaohu and Shou, Mike Zheng},
  booktitle=ICCV,
  year={2023}
}

@inproceedings{fatezero,
  title={{FateZero}: Fusing Attentions for Zero-shot Text-based Video Editing},
  author={Qi, Chenyang and Cun, Xiaodong and Zhang, Yong and Lei, Chenyang and Wang, Xintao and Shan, Ying and Chen, Qifeng},
  booktitle=ICCV,
  year={2023}
}

@article{anyv2v,
  title={{AnyV2V}: A Tuning-Free Framework for Any Video-to-Video Editing Tasks},
  author={Ku, Max and Wei, Cong and Ren, Weiming and Yang, Harry and Chen, Wenhu},
  journal={Transactions on Machine Learning Research},
  year={2024}
}

@article{stablev2v,
  title={{StableV2V}: Stabilizing Shape Consistency in Video-to-Video Editing},
  author={Liu, Chang and Li, Rui and Zhang, Kaidong and Lan, Yunwei and Liu, Dong},
  journal={IEEE Transactions on Circuits and Systems for Video Technology},
  volume={36},
  number={4},
  pages={5467--5482},
  year={2026}
}

@misc{gpt5,
  author={{OpenAI}},
  title={Introducing {GPT-5}},
  howpublished={\url{https://openai.com/index/introducing-gpt-5/}},
  year={2025}
}

@misc{gemini,
  author={{Google}},
  title={{Gemini 3}},
  howpublished={\url{https://gemini.google.com/}},
  year={2025}
}

@inproceedings{instructpix2pix,
  title={{InstructPix2Pix}: Learning to Follow Image Editing Instructions},
  author={Brooks, Tim and Holynski, Aleksander and Efros, Alexei A.},
  booktitle=CVPR,
  year={2023}
}

@article{instructedit,
  title={{InstructEdit}: Improving Automatic Masks for Diffusion-based Image Editing With User Instructions},
  author={Wang, Qian and Zhang, Biao and Birsak, Michael and Wonka, Peter},
  journal={arXiv preprint arXiv:2305.18047},
  year={2023}
}

@inproceedings{smartedit,
  title={{SmartEdit}: Exploring Complex Instruction-based Image Editing with Multimodal Large Language Models},
  author={Huang, Yuzhou and Xie, Liangbin and Wang, Xintao and Yuan, Ziyang and Cun, Xiaodong and Ge, Yixiao and Zhou, Jiantao and Dong, Chao and Huang, Rui and Zhang, Ruimao and Shan, Ying},
  booktitle=CVPR,
  year={2024}
}

@inproceedings{omnigen,
  title={{OmniGen}: Unified Image Generation},
  author={Xiao, Shitao and Wang, Yueze and Zhou, Junjie and Yuan, Huaying and Xing, Xingrun and Yan, Ruiran and Li, Chaofan and Wang, Shuting and Huang, Tiejun and Liu, Zheng},
  booktitle=CVPR,
  year={2025}
}

@article{step1x_edit,
  title={{Step1X-Edit}: A Practical Framework for General Image Editing},
  author={Liu, Shiyu and Han, Yucheng and Xing, Peng and Yin, Fukun and Wang, Rui and Cheng, Wei and Liao, Jiaqi and Wang, Yingming and Fu, Honghao and Han, Chunrui and Li, Guopeng and Peng, Yuang and Sun, Quan and Wu, Jingwei and Cai, Yan and Ge, Zheng and Ming, Ranchen and Xia, Lei and Zeng, Xianfang and Zhu, Yibo and Jiao, Binxing and Zhang, Xiangyu and Yu, Gang and Jiang, Daxin},
  journal={arXiv preprint arXiv:2504.17761},
  year={2025}
}

@article{bagel,
  title={Emerging Properties in Unified Multimodal Pretraining},
  author={Deng, Chaorui and Zhu, Deyao and Li, Kunchang and Gou, Chenhui and Li, Feng and Wang, Zeyu and Zhong, Shu and Yu, Weihao and Nie, Xiaonan and Song, Ziang and Shi, Guang and Fan, Haoqi},
  journal={arXiv preprint arXiv:2505.14683},
  year={2025}
}

@misc{qwen_image_edit,
  title={{Qwen-Image-Edit}: Image Editing with Higher Quality and Efficiency},
  author={{Qwen Team}},
  year={2025},
  howpublished={\url{https://qwen.ai/blog?id=qwen-image-edit}}
}

@misc{flux1_kontext,
  title={Introducing {FLUX.1 Kontext} and the {BFL} Playground},
  author={{Black Forest Labs}},
  year={2025},
  howpublished={\url{https://bfl.ai/blog/flux-1-kontext}}
}

@article{seedream4,
  title={{Seedream 4.0}: Toward Next-generation Multimodal Image Generation},
  author={{Team Seedream} and Chen, Yunpeng and Gao, Yu and Gong, Lixue and others},
  journal={arXiv preprint arXiv:2509.20427},
  year={2025}
}

@misc{nano_banana_2,
  title={Build with {Nano Banana 2}, Our Best Image Generation and Editing Model},
  author={Fortin, Alisa and Alessio, Bea},
  year={2026},
  howpublished={\url{https://blog.google/innovation-and-ai/technology/developers-tools/build-with-nano-banana-2/}}
}
}

\newpage
\section*{Appendix}

\subsection*{Formal Details of The Motion Encoder}
Let $U^{(\ell-1)}$ denote the input of the $\ell$-th layer. 
We first perform spatial attention independently within each frame:
\begin{equation}
\tilde{U}^{(\ell)}_{t,:,:} = U^{(\ell-1)}_{t,:,:} + 
\mathrm{Attn}_{\text{spatial}}\!\left(\mathrm{LN}(U^{(\ell-1)}_{t,:,:})\right),
\end{equation}
where $\mathrm{Attn}_{\text{spatial}}$ attends over all spatial locations $(h,w)$ at fixed time $t$.
We then perform temporal attention independently at each spatial location:
\begin{equation}
\bar{U}^{(\ell)}_{:,h,w} = \tilde{U}^{(\ell)}_{:,h,w} +
\mathrm{Attn}_{\text{temporal}}\!\left(\mathrm{LN}(\tilde{U}^{(\ell)}_{:,h,w})\right),
\end{equation}
where $\mathrm{Attn}_{\text{temporal}}$ attends over all timesteps at fixed spatial position $(h,w)$.
Finally, the output of the layer is obtained with a feed-forward network:
\begin{equation}
U^{(\ell)} = \bar{U}^{(\ell)} + \mathrm{MLP}\!\left(\mathrm{LN}(\bar{U}^{(\ell)})\right).
\end{equation}

After $L_m=6$ layers, the final motion representation is
\begin{equation}
h_{\text{mot}} = U^{(L_m)}.
\end{equation}
We inject $h_{\text{mot}}$ into the frozen Wan2.2 DiT through a ControlNet-style residual branch~\citep{controlnet}, allowing the generator to preserve source motion while leaving appearance and editing semantics to be determined by the edited anchors. The overall framework of the motion encoder is shown in Figure~\ref{fig:motion}.

\subsection*{Training Stage-2}

To train the lightweight motion branch, we keep the Wan 3D VAE and DiT frozen and optimize only $\mathcal{M}_{\theta}$ and the ControlNet-style residual branch.
Given a training target video $V^{\star}$ with latent code
\begin{equation}
Z^{\star} = \mathcal{V}_{3\text{D}}^{\text{enc}}(V^{\star}),
\end{equation}
we sample a diffusion timestep $\tau$ and noise $\epsilon$, and obtain a noisy latent video
\begin{equation}
X_{\tau}^{\star} = \alpha_{\tau} Z^{\star} + \sigma_{\tau}\epsilon.
\end{equation}
To mimic inference, the selected anchor frames are replaced by their clean latent values:
\begin{equation}
\tilde{X}_{\tau}^{\star}
=
M_{\mathcal{A}} \odot Z_{\mathcal{A}}^{\star}
+
(1-M_{\mathcal{A}})\odot X_{\tau}^{\star}.
\end{equation}
The model predicts the noise residual
\begin{equation}
\hat{\epsilon}
=
D_{\psi}(\tilde{X}_{\tau}^{\star}, h_{\text{mot}}),
\end{equation}
and we compute the training loss only on non-anchor frames:
\begin{equation}
\mathcal{L}_{\text{vid}}
=
\left\|
(1-M_{\mathcal{A}})\odot (\hat{\epsilon} - \epsilon)
\right\|_2^2.
\end{equation}
This objective explicitly teaches the model to interpolate and propagate content between fixed anchors, rather than re-synthesizing the anchors themselves.

\subsection*{Broader Impacts}
This work focuses on foundational research in video generation.
We did not identify any specific direct societal impacts beyond the general considerations that accompany generative media.

\end{document}